\documentclass[10pt, conference, compsocconf]{IEEEtran}
\ifCLASSINFOpdf

\else

\fi

\usepackage{amssymb}
\usepackage{graphicx}
\usepackage{times}
\usepackage{epsfig}
\usepackage{amsmath}
\usepackage{subfigure}
\usepackage{amssymb}
\usepackage{algorithm}
\usepackage{algorithmic}
\usepackage{epstopdf}
\usepackage{mathtools}

\begin{document}
\title{Patch-based Texture Synthesis for Image Inpainting}

\author{
    \IEEEauthorblockN{Tao Zhou, Brian Johnson, Member, \emph{IEEE}, Rui Li\IEEEauthorrefmark{2}}

}

\maketitle

\begin{abstract}
Image inpaiting is an important task in image processing and vision. In this paper, we develop a general method for patch-based image inpainting by synthesizing new textures from existing one. A novel framework is introduced to find several optimal candidate patches and generate a new texture patch in the process. We form it as an optimization problem that identifies the potential patches for synthesis from an coarse-to-fine manner. We use the texture descriptor as a clue in searching for matching patches from the known region. To ensure the structure faithful to the original image, a geometric constraint metric is formally defined that is applied directly to the patch synthesis procedure. We extensively conducted our experiments on a wide range of testing images on various scenarios and contents by arbitrarily specifying the target the regions for inference followed by using existing evaluation metrics to verify its texture coherency and structural consistency. Our results demonstrate the high accuracy and desirable output that can be potentially used for numerous applications: object removal, background subtraction, and image retrieval.
\end{abstract}

\IEEEpeerreviewmaketitle

\section{Introduction}
In image processing, inpainting is an essential task that uncover the potential textures in a missing region or target region by using the known information from the remaining region on the image. The goal is to make the result image as plausible as possible in terms of texture consistency by minimizing the artifacts. Image inpainting is also called as image completion that enables numerous applications in digital media from photo-editing including object removal, background recovery, to context-aware based image retrieval for visual data searching, to 3D reconstruction from multiple views. For all of these tasks, image inpainting plays as a initial step and benefit all of these subsequent vision tasks.

In literature, Various methods have been proposed to achieve the recovery of missing pixels. In \cite{01}, \cite{04}, \cite{05}, the missing pixels are interpolated using neighbor pixels based on the image properties such as smoothness of images. In \cite{03}\cite{06}\cite{07}, the image inpainting is achieved based pattern similarity. In \cite{03}, the authors present a novel and efficient algorithm that combines the advantages of these two approaches. In their paper, authors have used a data term priority which is determined in the direction of isophotes. The greedy search algorithm implemented in the above exemplar-based approaches is computationally intensive. This is addressed by Qiang Chen et.al \cite{04} by using a search strategy applied to a restricted range only. Driori et al. in their article \cite{09} incorporated a pyramid image approximation and adaptive image fragments to obtain satisfactory results. The drawback of these algorithms is that the diffusion process introduces some blur, which becomes noticeable when filling alrger regions. To retrieve this drawback, another kind of algorithms based on texture synthesis method was employed\cite{09}\cite{10}, but these algorithms consumes much computation time and also fail to reconstruct the feature of structural texture accurately.

In the texture matching period considering the efficiency, we propose an image inpainting, based on exemplar based image inpainting algorithm in successive elimination algorithm \cite {01}, our approach at least execution and previous techniques designed for recovery of fine scratches, and in which the object is removed, it outperforms in the aspects of computational efficiency early work.

\section{The Approach}
The core of exemplar-based image inpainting algorithm is an isophote-drived image- sampling process. Texture synthesis replicates both texture and structure information; while the structure propagation depends on the filling order which is determined by the priority. During the period of texture matching, successive elimination algorithm is employed to improve the computational efficiency. In this model, an energy functional is minimized to segment and denoise an image. The MS energy functional is given by

\begin{equation}
I_t = \frac{1}{k_p}\cdot\sum_{q\in \Omega}I_qf(||p - q||)\cdot g(||I_p - I_q||)
\end{equation}

Nevertheless, by minimizing a sum of squared differences on each pixel and applying the greedy Algorithm 1, we can not ensure the global minimization of the energy. In order to clarify the difference between minimizing the previous energy and applying Algorithm 1, let us give a simple example. Suppose that we are trying to inpaint an image containing repetitive patterns. An example of such an image is shown  and the results obtained after applying Algorithm 1 with two different patch sizes are shown on figures 1 and
2. As it can be noticed the result is perfect with $5\times 5$ patches but not with $9\times 9$ patches. Since this image contains repetitive patterns, we know that the global minimum of the Energy (1) is equal to 0 because the gap $\Omega$ can be filled by verbatim copy of a piece of $\Omega^c$
, with the same shape and size as  $\Omega$. If we compute the value of the energy (1) after Algorithm 1 has been
applied, we see that for $9\times 9$  patches it is indeed equal to 0, while for $5\times 5$ patches it returns a much larger value.

Recast Eq. 1 in terms of the level set function $\Phi$ then
minimize the functional energy E with respect to $\mu1$,
$\mu2$, and $\Phi$, we obtain the equations for $\mu1$, $\mu2$, and $\Phi$:
\begin{equation}
\lambda(x, y)(\mu_1 - \mu_0) = \mu\Delta^2\mu1 \mbox{, where } \frac{\delta\mu_1}{\delta n} = 0
\end{equation}

\begin{equation}
\lambda(x, y)(\mu_2 - \mu_0) = \mu\Delta^2\mu2 \mbox{, where } \frac{\delta\mu_2}{\delta n} = 0
\end{equation}

The smooth image functions $\mu1$ and $\mu2$ can be obtained
by solving the damped Poisson Eqs. (1) and (2), and the segmentation curve can be inferred according
to Eq. (3). This is a piecewise smooth approximation proposed by Chan and Vese. Many advantages can be achieved by this method, for example by using multiple level set functions for simultaneous segmentation and smoothing of noisy images, and the detection of three knots. However, due to the three partial differential equations to be solved simultaneously, the computational cost of this method is very large. To overcome this difficulty, Chan and Vese are proposed by using piecewise constant approximation another approximation method. If the intensity of the image in different regions is uniform, the intensity of the image in different regions can be approximated by a constant. In this case, the MS
energy function can be simplified to the equation (4):

\begin{equation}
E(c_k, C) = \sum_k\int_{\Omega_k} \lambda(x, y)(c_k - \mu_0)^2dxdy + V|C|
\end{equation}
where $\Omega_k$ represents the area inside each region. The
gradient term in the MS energy functional disappears
in Eq. (2) because the gradient inside each region is
zero. Using the level set method \cite{02} and the MS
energy functional of the two phase segmentation, the
image is segmented into two regions. To minimize
the energy functional with respect to $c_1$, $c_2$, and $\Phi$,
we obtain the following equations:
\begin{equation}
\frac{\delta\Phi}{\delta t} = \delta(\Phi) [\lambda(x, y)(\mu_0 - c_1)^2 - \lambda(x, y)(\mu_0 - c_2)^2]
\end{equation}

With the level set method, MS model is very good in many applications. However, generally only one or two level set functions are used to segment the image for two or four stages. Therefore, most of the experiments have been completed simple image. For images with complex structure, area is not in the image by one or two level set function. Generally, we use more neurons, separate the plurality of layers, and the number of neurons is adjustable. However, when the number of neurons is so close to 255, there may be some never excite neurons.
Through experiments, we found that, on the one hand, if there is too little separation layer, the final repair results will be very rough. On the other hand, if there are too many separate layers, not only the damaged pixels may be difficult to determine which layer they belong to, and lead to obvious patch errors, but also consider a lot of time for calculation.
Suppose we construct a SOM having $m\times n$ n neurons to
separate an image into $m\times n$ n different layers. And each
neuron has 3 input knobs, for each pixel having three color
(Red, Green and Blue) intensity

\section{Experiment Results}

As shown in Fig.1 of Objects removal, Fig. 1(c) and Fig.
1(b) respectively show the results of the improved algorithm
and the algorithm from \cite{01} based on OpenCV3.0 (Running
environment: Core2 Q4600-2.4GHz processor, 2G memory),
and the results denote that the images inpainted by these two
algorithm are visually same, but the computational cost of
the our improved algorithm is just 110 seconds while the
inpainted algorithm from \cite{01} is 233 seconds. In the second
time of object removal, the computational cost of these two
inpainted algorithms are respectively 22 seconds and 43
seconds. Fig. 2 shows the result of single object removal, the
resolution of the original image is $400\times 300$, the
computational cost of these two inpainted algorithms are
respectively 32 seconds and 82 seconds.

\begin{figure}[htb]
\centerline{\epsfig{figure=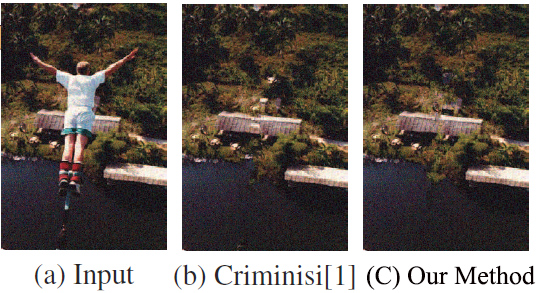, width=9cm}}
\caption{Image inpainting results comparison}\label{fig:evalue1}
\end{figure}

A0 and B0 have the same pixel intensity of the image in the reference. According to the photography reciprocity, when change the exposure time, A0 and B0 change the pixel value of the corresponding. Intuitively, can be reproduced from the image degradation value of A B in the absence of image capture, sensor noise, which is usually caused by the process of sampling noise and produce compression noise. Therefore, all of the pixels are more accurately with the same strength.

\begin{figure}[htb]
\centerline{\epsfig{figure=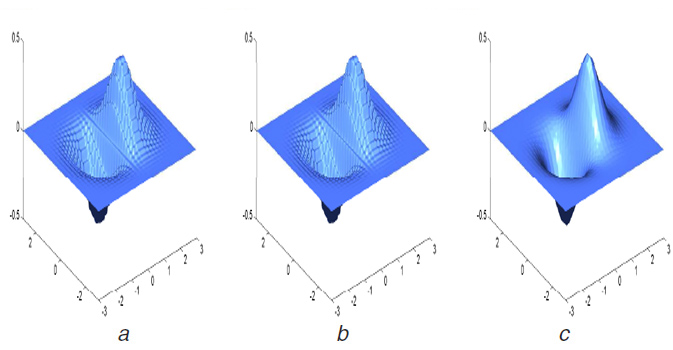, width=9cm}}
\caption{Error estimation curves: initial values; after smoothing with median filter; after extending empty values at end}\label{fig:evalue1}
\end{figure}

Finally, we demonstrate the results of using the same images as  TV
inpainting scheme and our inpainting method for a image with the size of
 $400\times 300$ pixels photo, which is shown in Figure 3
as the original image. The photo has been preprocessed. In
the process, the color intensity which equal to 255 has been
change to 254, so that we can use the 255th color intensity to
indicate the damaged pixels. The damaged result of the
photo is shown figure 3. The repair scheme of the TV image result is we get repair algorithm optimization based on patch finder.

\begin{figure}[htb]
\centerline{\epsfig{figure=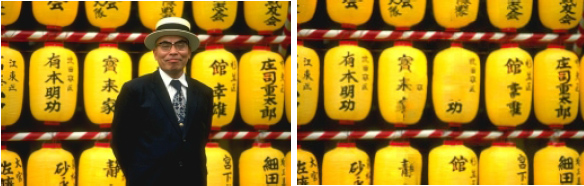, width=9cm}}
\vspace{10mm}
\centerline{\epsfig{figure=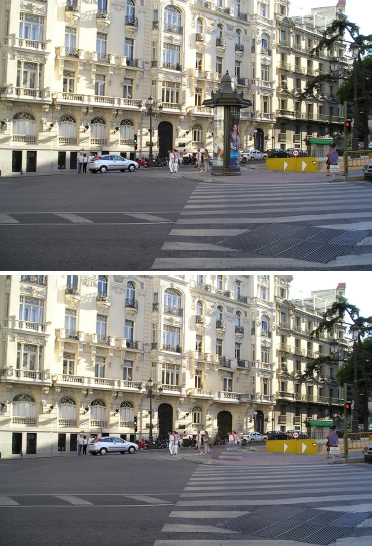, width=9cm}}
\caption{More results on Image inpainting}\label{fig:evalue1}
\end{figure}

\section{Conclusion}
This paper presents an improved algorithm remove objects and repair damaged image. As a result, under the selected object it has been possible in the case of the visual appearance of the image to be replaced, mimicking the source area. Our method based on a unified approach to texture synthesis technique used to determine the model of filling a modulation sequence of the target area. Keeping with the pixel confidence value illuminance image, the impact of their priority to fill. To improve efficiency, the gradual elimination algorithm to obtain the global optimal solution.

\end{document}